\documentclass{article}

\usepackage{arxiv}

\usepackage[utf8]{inputenc} 
\usepackage[T1]{fontenc}    
\usepackage{hyperref}       
\usepackage{url}            
\usepackage{booktabs}       
\usepackage{amsfonts}       
\usepackage{nicefrac}       
\usepackage{microtype}      
\usepackage{lipsum}		
\usepackage{graphicx}
\usepackage[super,sort]{natbib}
\usepackage{doi}
\usepackage{float}
\usepackage{booktabs}
\usepackage{tabularx}
\usepackage{array}
\usepackage{makecell}

\title{Faces or Models? Where the Age Gap in Facial Expression Recognition Comes From}

\author{ \href{https://orcid.org/0000-0000-0000-0000}{\includegraphics[scale=0.06]{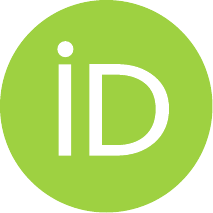}\hspace{1mm}Jessica Wang}\\
	Trinity School NYC\\
New York, NY 10024, USA
}

\date{~}

\hypersetup{
pdftitle={A template for the arxiv style},
pdfsubject={q-bio.NC, q-bio.QM},
pdfauthor={David S.~Hippocampus, Elias D.~Striatum},
pdfkeywords={First keyword, Second keyword, More},
}

\begin{document}
\maketitle

\begin{abstract}
Facial expression recognition (FER) systems are increasingly used in elder care, yet their accuracy on older faces is rarely measured. This study evaluated five open-source FER models on the FACES database, which holds 2,052 posed photographs of 171 adults aged 19 to 80\cite{ref1}. The models cross three architectures (ResNet-18, POSTER++ and QCS) with two training datasets (RAF-DB and AffectNet)\cite{ref1}. All models lost accuracy on older faces by 11.7 to 15.2 percentage points. Older faces were harder for people too: human raters lost 8.6 points, a classifier using only facial geometry lost 8.3, and for anger, fear, and happiness older adults differed more from one another when posing. On specific expressions the models fell much further than people did. AffectNet-trained models misread older neutral faces as sad or angry 62\% of the time. RAF-DB-trained models lost fear, disgust, and anger instead. The gap persisted after controlling for architecture, image quality, cropping, and expression intensity. When only the final classification layer was refitted on older faces, the gap shrank from 11 to 21 points to 4 points or less, suggesting that the models' learned features can represent older faces and that the deficit lies in how the final layer was trained.
\end{abstract}


\raggedbottom
\section{Introduction}
About 1.1 billion people worldwide are aged 60 or older, and this number is projected to reach 2.1 billion by 2050\cite{ref2}. Many of them face health conditions that require regular monitoring: around 15\% of adults aged 70 and older live with a mental disorder, such as anxiety or depression\cite{ref2}. Facial expressions offer a non-invasive window onto how a person is feeling. Automated facial expression recognition (FER) is being studied as a way to detect early signs of low mood and cognitive decline in older adults\cite{ref3,ref4,ref5}.

However, these systems are trained mainly on internet photographs of younger adults\cite{ref6}. Older faces differ: their skin is more lined and less taut, lighting and image quality can be different, and older adults may move the face differently to show the same feeling\cite{ref7}. Whether a model trained on younger faces reads older ones just as well is rarely measured. This study asks how accurately five publicly released FER models read the expressions of older adults, and where their errors come from.

Older adults are the people these systems are pointed at. The population is aging rapidly, and older adults are more prone to conditions that call for sustained monitoring and specialized care\cite{ref2}. Automated expression analysis is being introduced in the settings that older adults are part of, from care homes to clinical monitoring and companion robots. In each of these, an older adult is more likely to be the subject of the system than its operator, and thus less able to notice a misreading, question it, or opt out.

This measurement gap can turn into a practical harm. The results reported here show that all five models read older faces less accurately than younger ones. This applies most of all to neutral faces, which a resting face displays most of the time. This finding agrees with Kim et al., who found that four commercial FER systems were least accurate on older adults\cite{ref6}. A system that reads a resting older face as sad or angry does not fail quietly: it reports distress that is not there, in settings where a flagged emotional state may inform a clinical judgement or a care decision\cite{ref6}.

This study evaluates how accurately open-source FER models classify the static facial expressions of older adults, defined here as participants aged 69 and older who are labelled `old' in the FACES database\cite{ref1}. Five released models were chosen: QCS trained on RAF-DB, QCS trained on AffectNet-7, POSTER++ trained on RAF-DB, POSTER++ trained on AffectNet-7, and NLA trained on RAF-DB. Each takes a different approach to the task. QCS refines facial features by comparing images, POSTER++ combines facial appearance with landmark structure through cross-attention, and NLA addresses ambiguous and noisy training labels. As two architectures appear with both training datasets, the design separates the effect of the architecture from the effect of the data it was trained on.

Beyond measuring the age gap, the study asks where it comes from. It separates the part that lies in the faces themselves, measured without any model through human ratings and a geometry-only classifier, from the part that lies in the models, probed by refitting only each model's final layer. It then tests the explanations that have been proposed for the gap, including image quality, cropping, expression intensity, model architecture and training dataset.

Facial expression recognition for older adults has drawn comparatively little work. The most recent systematic review of the area found only 31 deep-learning FER studies on elderly populations across a full decade, from 2015 to September 2024, with 285 records retrieved across five databases\cite{ref8}. Older adults were less represented in the data that these models learn from. Widely used corpora, such as AffectNet, carry no age labels and hold few older faces\cite{ref9}. The deficit is therefore partly invisible by construction: a model cannot be evaluated by age on a dataset that does not record it, so it can report strong overall accuracy while doing much worse on older faces without that ever surfacing. Corpora built under controlled conditions with deliberate age diversity, such as FACES, remain the exception\cite{ref1}.

Kim et al. benchmarked four commercial FER systems on the same FACES corpus used here, across the same three age groups, and found that all four did best on younger and worst on older adults\cite{ref6}. That study established the disparity, but it did not explain it. This is the wider pattern: of the 31 studies in the review above, only six examined the effect of age on performance at all\cite{ref8}.

Furthermore, in studies where causes have been proposed, they have generally been asserted or tested one at a time rather than against each other. Three recur: age-related wrinkles and folds that alter how an expression reads\cite{ref10}; a decline in facial-muscle function with age\cite{ref7}; and under-representation of older faces in training data, inferred from a 22\% accuracy gain that Park et al. obtained by training separately on children's and adults' faces\cite{ref11}. The first two are already in tension, as Grondhuis et al. argued for muscle function over wrinkles\cite{ref7}, and none of the three has been separated from the others on a shared corpus with a shared set of models.

This study makes two unique contributions. First, it evaluates five released models on older adults in a design that crosses architecture with training dataset, so a shared failure cannot be blamed on one model family or one dataset. Second, it separates the age gap into a part that comes from the faces and a part that comes from the models, and it tests each proposed cause directly rather than by inference. The wrinkle explanation, for example, is contradicted by the results here.

\section{Methods}
\label{sec:headings}
\subsection{Research design}

This study is an observational, cross-sectional evaluation of five released FER checkpoints applied, without fine-tuning, to a controlled database of posed facial expressions spanning three age groups. The design crosses three architectures with two training datasets so that the effect of architecture can be separated from the effect of training data. This section first describes the three architectures and the two training datasets, then the FACES database on which the models are tested, and finally the preprocessing, scoring and analysis procedures.

\subsection{Models evaluated: QCS}
Wang et al. designed Quadruplet Cross Similarity (QCS), an open-source model released on GitHub\cite{ref12}. It aims to stop the network from learning interfering features, cues that happen to correlate with an expression without being part of it. For example, when a model learns to spot happiness, raised mouth corners and defined cheeks are genuine cues, but the lighting or skin tone common to happy training photographs are not, and a model that leans on them will generalize poorly.

QCS uses IR-50, a network pre-trained for face recognition on MS-Celeb-1M, as its backbone\cite{ref13}. Its main innovation is Cross Similarity Attention, a spatial attention mechanism that decides where on the face to look. The standard attention used in Vision Transformers compares two images with a dot product. Wang et al. pinpointed that when two different images of the same expression are compared this way, matching regions respond only weakly, because the dot product mixes the magnitude and the direction of the feature vectors\cite{ref12}. Cross Similarity Attention instead measures similarity with the Euclidean, or straight-line, distance between image features, which responds strongly where two images match.

QCS trains on four images at a time: an anchor, a second image of the same expression, and two images of a different expression. A three-image scheme would pull the anchor toward the matching image and away from the differing one at once, and those opposing updates can conflict. The fourth image makes the arrangement symmetric and steadies training. What the four-image comparison learns is then passed back into a single branch through a step the authors call contrastive residual distillation. The trained model therefore classifies one image at a time, as a real application would. On the standard benchmarks, QCS reached 92.5\% accuracy on RAF-DB and 67.94\% on AffectNet-7, ahead of the other models in the authors' comparison\cite{ref12}.

\subsection{Models evaluated: POSTER++}
POSTER++ is an open-source model built as a simpler, cheaper successor to POSTER\cite{ref14}. Rather than relying on facial appearance alone, it combines two streams: appearance features from an IR-50 backbone and structural information from facial landmarks, processed with MobileFaceNet. Landmarks locate the eyes, eyebrows, nose and mouth, and POSTER++ fuses the two streams with cross-attention so the model can relate how the face looks to how it is built.

Two changes make it efficient. A multi-scale mechanism folds facial information at several levels of detail directly into the features, in place of POSTER's heavier pyramid design. Cross-attention runs within local windows of the feature map rather than across the whole of it, which cuts the cost of the operation while still letting landmark and appearance information interact. Mao et al. reported accuracies of 92.21\% on RAF-DB and 67.49\% on AffectNet-7\cite{ref14}. In this study, the constructed network measured 58.3 million parameters, larger than the figure the paper reports; parameter counts here were therefore taken from the networks as built rather than from the literature.

\subsection{Models evaluated: NLA}
Navigating Label Ambiguity (NLA) applies a new training scheme to a standard ResNet-18\cite{ref15}. It targets two problems that are common in FER datasets collected in the wild: label ambiguity, where annotators disagree on an expression or a face is mislabeled, and class imbalance, where some expressions appear far more often than others. NLA handles these with Noise-aware Adaptive Weighting, which raises the weight of clearly labelled examples and lowers the weight of doubtful ones, and adjusts those weights as training proceeds and the model learns which features matter. It also scores each image alongside its mirror image and encourages the two to receive the same label, which makes the weighting steadier.

NLA reached 89.93\% on RAF-DB and 67.06\% on AffectNet-7, ahead of the previous best models on both\cite{ref15}. Its edge is largest on deliberately corrupted data: when Lee et al. added mislabeled images or forced class imbalance, NLA outperformed its competitors at every level, so it is suited to noisy, imbalanced real-world data rather than clean benchmarks\cite{ref15}.

\subsection{Training datasets}
QCS, POSTER++ and NLA were each evaluated in the versions trained on RAF-DB or on AffectNet-7. RAF-DB holds about 30,000 facial images collected from the internet and labelled by around 40 annotators each, whose labels were then reconciled\cite{ref16}. Its single-label subset contains 15,339 images across the seven basic classes (anger, disgust, fear, happiness, neutrality, sadness and surprise), and this is the subset most FER research uses. AffectNet-7 is a seven-class subset of AffectNet, which gathers more than a million internet images, about 450,000 of them hand-labelled\cite{ref9}. AffectNet-7 uses roughly 284,000 images for training and is markedly imbalanced, with far more happy and neutral faces than disgusted or fearful ones, which makes it a useful test of whether a model can recognize the rarer expressions. Neither dataset was built for age balance.

The models are tested not on RAF-DB or AffectNet but on FACES, a database built under controlled studio conditions across three age groups\cite{ref1}. Because only the face changes from one photograph to the next, FACES is well suited to studying how recognition accuracy varies with age.

\subsection{Participants and data}
\begin{figure}[H]
	\centering
	\includegraphics[scale=1]{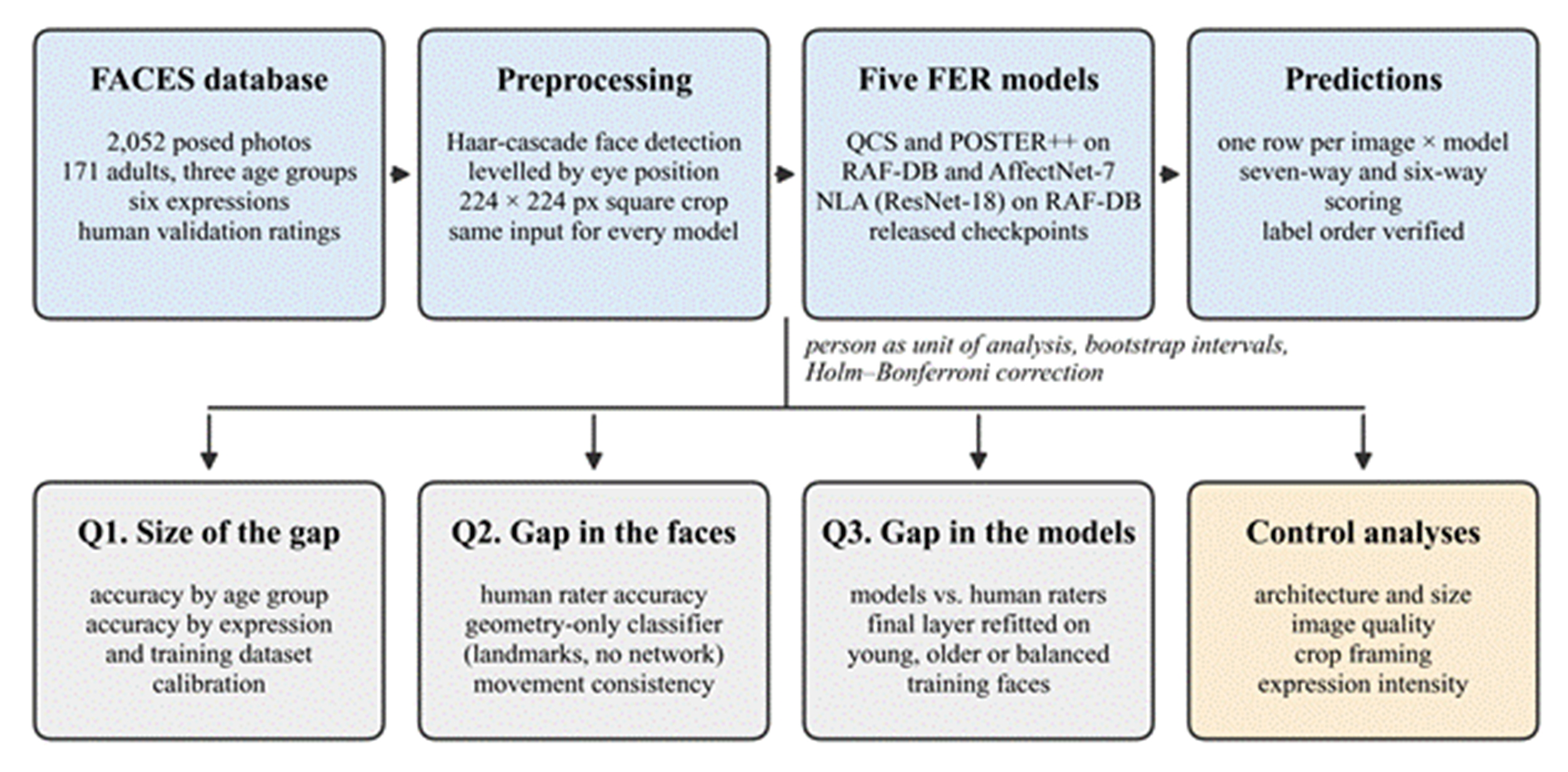}
	\caption{Overview of the study design. Every model scores the same cached crops.}
	\label{fig:fig1}
\end{figure}
The FACES database contains 2,052 posed color photographs of 171 adults in three age groups: young (19--31 years, n = 58), middle-aged (39--55, n = 56) and older (69--80, n = 57)\cite{ref1}. Each person posed six expressions (anger, disgust, fear, happiness, neutrality and sadness) in two separate photo sets, giving 12 photographs per person. Lighting, pose and background are the same in every photograph, and each age-by-expression cell contains 112 to 116 images. Labels were read from the file names, and the cell counts were checked before any model was run.

In addition, FACES provides validation ratings collected from 154 raters of three ages\cite{ref1}. For each photograph, the ratings give the percentage of raters who chose the intended expression from the same six options. This study uses this percentage as human accuracy.

\subsection{Preprocessing}

Image preprocessing (face detection, alignment, cropping and resizing) was carried out with the agreement of the FACES Technical Agent. Each image was cropped once and cached, so every model received identical input. Faces were detected with an OpenCV Haar cascade, levelled using the positions of the eyes, cropped to a square and resized to 224 $\times$ 224 pixels. Detection succeeded on all 2,052 images. The detected boxes were about 9\% larger on older faces (median 2,031 compared with 1,859 pixels). The crop margin was fixed at zero, a tight crop like RAF-DB's training images, because the RAF-DB models lose accuracy quickly when the crop is looser; this is explained in detail in the Results.

\subsection{Variables and measurements: models and scoring}
Five released checkpoints were evaluated without fine-tuning (Table~\ref{tab:tab1}). QCS and POSTER++ were each tested with both training datasets and NLA with RAF-DB, so architecture and training data vary independently. All three architectures start from backbones pre-trained for face recognition on MS-Celeb-1M\cite{ref13}, and POSTER++ and QCS load an identical backbone checkpoint. Each repository's own test-time preprocessing was reproduced. The order of each model's output labels was checked empirically, and happiness recall was at least 0.99 for every model under the verified order.
\begin{table}[H]
    \caption{Models evaluated. Parameter counts were measured from the constructed networks.}
    \centering
    \begin{tabular}{lcc}
        \toprule
        \textbf{Architecture (parameters)} & \textbf{RAF-DB} & \textbf{AffectNet-7} \\
        \midrule
        QCS (82.9M)                 & Yes & Yes \\
        POSTER++ (58.3M)            & Yes & Yes \\
        NLA / ResNet-18 (11.7M)     & Yes & No  \\
        \bottomrule
    \end{tabular}
    \label{tab:tab1}
\end{table}

FACES contains six expressions, whereas the models output seven, including surprise. The primary score counts a surprise prediction as an error, as a deployed system would. Comparisons with human raters, who chose among six options, use a six-way score that ignores the surprise output. Because the FACES rating file offers raters only six response options, a seven-way score would penalize the models for offering a response, surprise, that human raters could never give; the six-way score removes that asymmetry and is used only for this comparison, not elsewhere in the paper.

\subsection{Data analysis}

The analyses answer three questions: how big the gap is, how much of it comes from the faces, and how much of it comes from the models, followed by control checks.

To measure the size of the gap, the study compared accuracy on younger and older faces for each model and each expression. It also measured calibration, the match between a model's confidence and its accuracy, using expected calibration error, and the agreement between models with Fleiss' $\kappa$.

To test whether part of the gap comes from the faces, two measures that involved no FER model were used. The first is human accuracy on younger and older faces. The second is a linear discriminant classifier trained on facial geometry alone. Faces were landmarked with a face-alignment network that locates 68 points, which was about equally confident on young and older faces\cite{ref17}. Each expressive photograph was aligned to the same person's neutral photograph with a similarity transform fitted on rigid landmarks (nose bridge, nose base and eye corners). It was then coded as the displacement of the remaining points, in units of the distance between the eyes, which removes identity and head pose.

The classifier separated the five emotional expressions. It was trained and tested within each age group, and it was also trained on young faces and tested on older ones. Each setting used 36 training people, different people for testing and 20 random splits. A third measure was how much people within each age group differed from one another in how they moved the face for each expression. Each person's movement was averaged over their two photo sets, and the difference between the two sets was used to remove landmark error (the detector's own imprecision in placing a point on a single image) and repeat-to-repeat variation (the real difference between how the same person posed the same expression on their two separate photographs) from the spread between people. A person's two neutral photographs gave a separate estimate of landmark noise.

To locate the rest of the gap in the models, the study first compared them with human raters on the six-way score. Then each network was frozen and a new linear classifier was trained on its learned features, using FACES photographs of 36 young people, 36 older people, or 18 of each. The strength of regularization was chosen by cross-validation within each training set. All three versions were tested on the same held-out older people, and on the held-out young people, over 20 random splits that used the same partition for every version to ensure that any difference between the three versions could only come from who was in their training data, not from an easier or harder test set. Features and sample size are fixed, so the versions differ only in the age of the training faces.

Four control analyses tested other explanations. Architecture was tested by comparing the three models trained on RAF-DB, which range from 11.7M to 82.9M parameters, and by contrasting architecture and training dataset across the four QCS and POSTER++ models. To check image quality, the age effect was adjusted for lighting and exposure (mean luminance, contrast, entropy, colorfulness and face size) in a logistic regression that treats each person's photographs as a cluster. Young and older faces were also compared within thirds of each quality measure to rule out image quality itself as a confound between the age groups. Crop framing was varied by re-cropping every face at margins of 0.0, 0.1, 0.2 and 0.3 of the detected box and scoring all five models again. Expression intensity was measured without a model; measuring intensity this way avoids circularity, since using one of the five models being evaluated would risk detecting that model's own biases rather than a genuine property of the expression. For six facial regions (brows, eyes, both cheeks, nasolabial region and mouth), the mean absolute pixel change was taken from the same person's neutral photograph and divided by the change in forehead and jaw, so that skin texture cannot inflate it. Error rates were then compared within six equal-sized intensity bins.

\subsection{Statistical analysis}
Each person in the FACES dataset contributes 12 correlated photographs, so the person is the unit of analysis. Confidence intervals are 95\% percentile bootstrap intervals from at least 2,000 resamples of whole people. Model comparisons are paired because every model scored the same images. Age gaps are reported in percentage points. Some analyses repeat over random splits of the same people, as in the refitting experiment and the geometry classifier. For these, each held-out person's accuracy is averaged over the splits in which they were held out, and the intervals and permutation p-values come from those person-level values. Tests over the splits themselves were avoided because the splits reuse the same people and overstate certainty: each of the roughly 57 people reappears across many of the 20 splits, so the results are not independent of one another the way a significance test assumes. In a simulation with no true effect, a Wilcoxon test over the splits gave a false positive in 25\% of datasets, five times the nominal 5\% rate a valid test is supposed to guarantee, and the person-level test in 3\%, confirming that the person-level test, not the per-split one, is the one the paper's other findings rely on. When several tests address one question, their p-values are adjusted with the Holm--Bonferroni method\cite{ref18}. This method ranks the p-values from smallest to largest and tests each against a progressively less strict threshold\cite{ref18}. It keeps the chance of any false positive within the group at 5\% and is less conservative than a plain Bonferroni correction.

\subsection{Ethical considerations}

This study analyzed the FACES database, a de-identified set of facial photographs released by its creators for research use under a data-use agreement, and did not involve any new data collection from human participants. No new interaction with human or animal subjects took place, so no new institutional review board approval was required; the FACES stimuli and their validation ratings were used in accordance with the database's terms of use.
\subsection{Data Availability}
FACES images and data derived from individual images cannot be shared by the author. Researchers may request access to the FACES database from the Max Planck Institute for Human Development (facesdatabase@mpib-berlin.mpg.de).
\section{Results}
\subsection{The age gap}
\begin{figure}[H]
	\centering
	\includegraphics[scale=0.7]{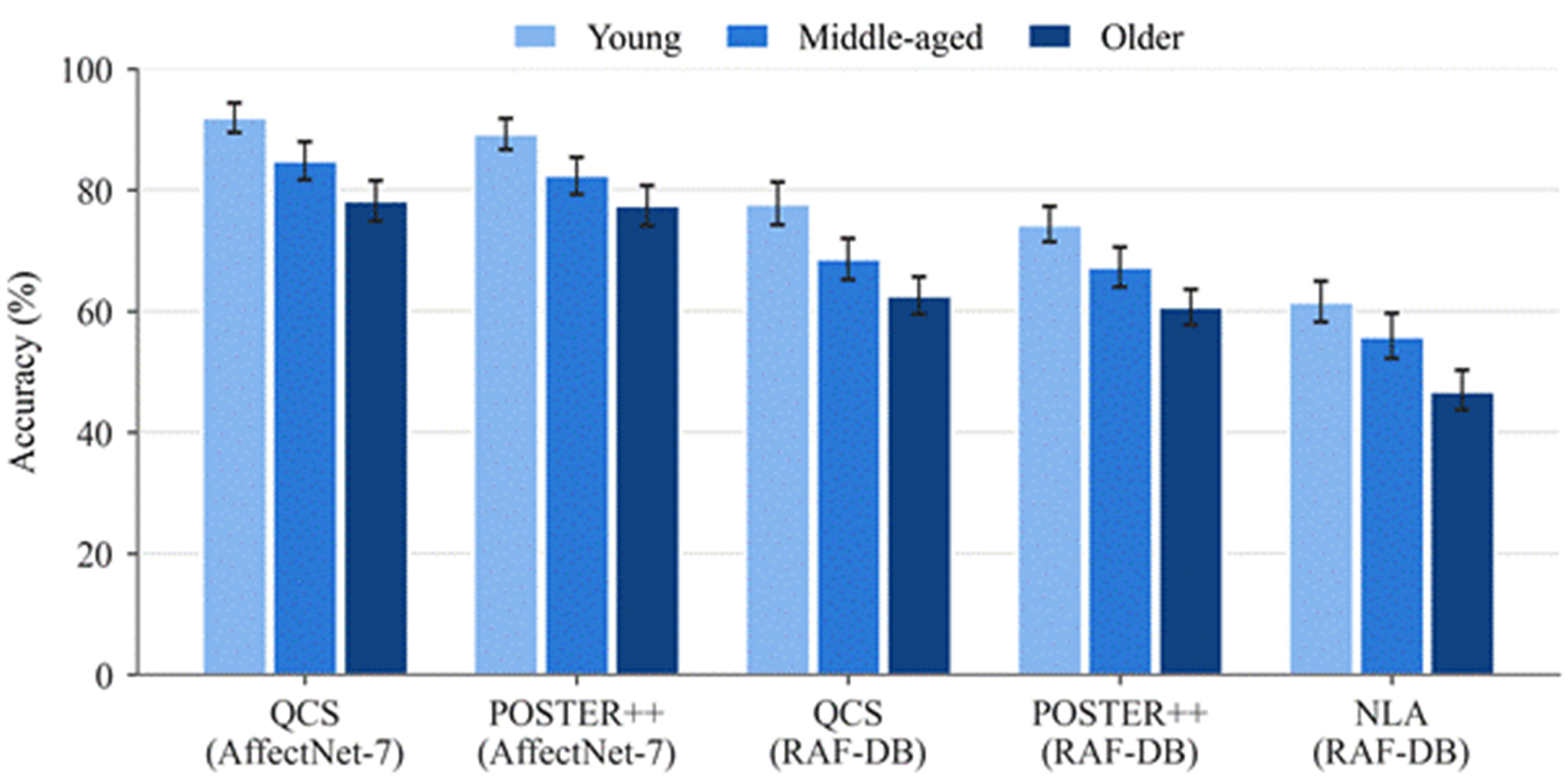}
	\caption{Accuracy by age group for each model. Error bars show the 95\% subject-bootstrap intervals Table \ref{tab:tab2}.}
	\label{fig:accuracy-age}
\end{figure}

Every model was less accurate on older faces, and accuracy fell steadily from young to middle-aged to older faces (Table \ref{tab:tab2}, Figure \ref{fig:accuracy-age}).
\begin{table}[H]
    \caption{Accuracy (\%) by age group with 95\% subject-bootstrap intervals, and the age gap (older minus young, percentage points) with its interval.}
    \centering
    \small
  \begin{tabular}{lccccc}
    \toprule
    \textbf{Model} &
    \textbf{Training data} &
    \textbf{Young} &
    \textbf{Middle-aged} &
    \textbf{Older} &
    \textbf{Gap (older--young)} \\
    \midrule

    QCS & AffectNet-7 &
    92.0 {\scriptsize [89.5, 94.4]} &
    84.8 {\scriptsize [81.7, 87.9]} &
    78.2 {\scriptsize [74.9, 81.6]} &
    -13.7 {\scriptsize [-18.1, -9.5]} \\

    POSTER++ & AffectNet-7 &
    89.2 {\scriptsize [86.6, 91.8]} &
    82.4 {\scriptsize [79.3, 85.4]} &
    77.5 {\scriptsize [74.1, 80.7]} &
    -11.7 {\scriptsize [-16.1, -7.4]} \\

    QCS & RAF-DB &
    77.7 {\scriptsize [74.3, 81.3]} &
    68.6 {\scriptsize [65.2, 72.0]} &
    62.6 {\scriptsize [59.5, 65.6]} &
    -15.2 {\scriptsize [-19.7, -10.6]} \\

    POSTER++ & RAF-DB &
    74.3 {\scriptsize [71.4, 77.3]} &
    67.3 {\scriptsize [64.0, 70.5]} &
    60.7 {\scriptsize [57.7, 63.6]} &
    -13.6 {\scriptsize [-17.7, -9.6]} \\

    NLA & RAF-DB &
    61.5 {\scriptsize [58.2, 64.9]} &
    55.8 {\scriptsize [52.2, 59.7]} &
    46.8 {\scriptsize [43.7, 50.3]} &
    -14.7 {\scriptsize [-19.4, -9.9]} \\

    \bottomrule
\end{tabular}
    \label{tab:tab2}
\end{table}

All five models decline steadily across the three age groups, with no group skipped or reversed. The two AffectNet-7 models stay the most accurate at every age (QCS 92.0\% then 84.8\% then 78.2\%; POSTER++ 89.2\% then 82.4\% then 77.5\%, young to middle-aged to older), while the three RAF-DB models are more spread out, from QCS's 77.7\% then 68.6\% then 62.6\% down to NLA's 61.5\% then 55.8\% then 46.8\%. Because the RAF-DB models start from a lower baseline, the same 12- to 15-point absolute drop in Figure~\ref{fig:accuracy-age} is proportionally steeper for them: NLA loses close to a quarter of its starting accuracy by older age, against about one-seventh for the two AffectNet-7 models.

The age gap ranged from 11.7 to 15.2 points, and every gap's 95\% interval excluded zero. The AffectNet models were the most accurate on older adults (78.2\% and 77.5\%), although their gaps (13.7 and 11.7 points) were similar to the RAF-DB models' (13.6 to 15.2). Being more accurate overall therefore did not make the AffectNet models any more resistant to the age-related drop itself.

\begin{figure}[H]
	\centering
	\includegraphics[scale=0.9]{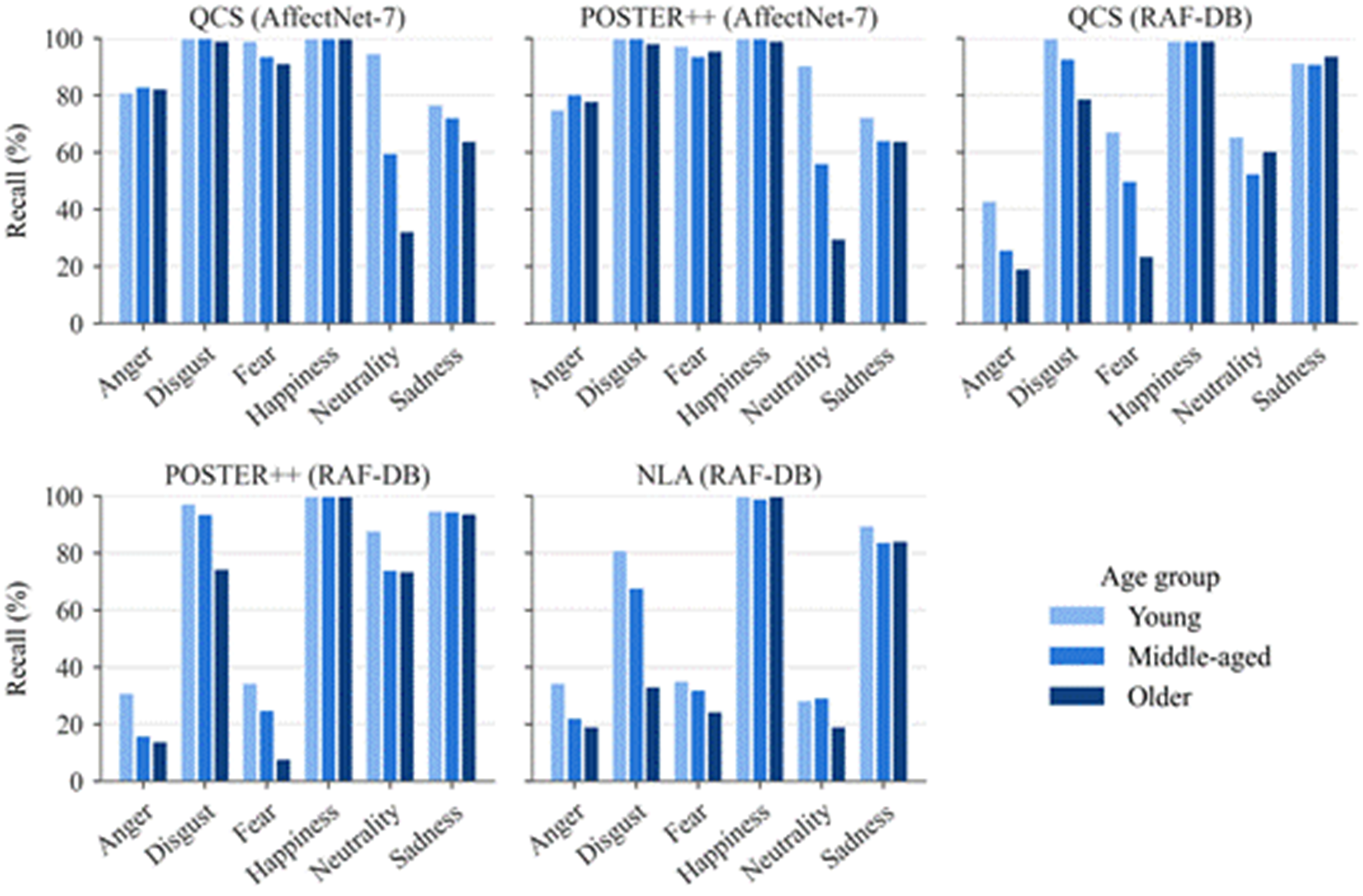}
	\caption{Recall for each expression by age group and model, on the seven-way score (a surprise response counts as an error). Table \ref{tab:expression-gap} uses the six-way score, which differs from this one mainly for fear.}
	\label{fig:fig3}
\end{figure}

The failures differed by training dataset (Figure~\ref{fig:fig3}). For the AffectNet models, recall on neutral faces fell from 90.5\% (POSTER++) and 94.8\% (QCS) for young adults to 29.8\% and 32.5\% for older adults. Most of the missed faces were read as sad or angry, about equally often (POSTER++ 34\% sad and 29\% angry; QCS 30\% sad and 31\% angry). Their other expressions changed by 13 points or less. For the RAF-DB models, the largest losses were for disgust, fear and anger. Neutral recall fell by 5 points for QCS, 9 for NLA and 14 for POSTER++. Happiness was recognized almost perfectly at every age.

Calibration worsened with age for the three RAF-DB models, which matters beyond raw accuracy: a system that grows both less accurate and less able to signal its own uncertainty on older faces is the more concerning failure mode for the deployment settings this paper is concerned with. Their expected calibration error rose from 0.19 (POSTER++), 0.20 (NLA) and 0.10 (QCS) on young faces to 0.30, 0.33 and 0.18 on older faces, whereas both AffectNet models stayed between 0.03 and 0.06 at every age. The models that lost the most accuracy on older adults were also the ones whose confidence scores were least trustworthy for that age group. Agreement between the models fell with age as well (Fleiss' $\kappa$ 0.69 for young, 0.61 for middle-aged and 0.54 for older faces).

\subsection{Part of the gap lies in the faces}
Human raters for the FACES dataset were less accurate on older faces (Table~\ref{tab:expression-gap}). Their accuracy fell from 85\% on young faces to 76\% on older faces, a gap of 8.6 points (95\% CI 6.7 to 10.5). Raters of every age showed a similar drop of 8 to 9 points, so older raters had no advantage on older faces.

The geometry-only classifier also lost accuracy on older faces. Trained and tested within each age group, its accuracy fell from 87.1\% on young faces to 78.8\% on older faces, a gap of 8.3 points (95\% CI 4.5 to 12.0; permutation p = 0.0001), against a chance level of 20\%. Trained on young faces and tested on older ones, it reached 75.9\%, only 2.9 points below the classifier trained on older faces (95\% CI 0.1 to 5.7; p = 0.05). A rule learned on young faces therefore transferred to older faces with a small further loss.

The landmark data suggest part of the reason. Landmark noise was similar at every age, since a person's two neutral photographs differed by about the same amount in each group (p = 0.09). Older adults did repeat each expression a little less closely across their two photo sets. For anger, disgust, fear and happiness, the average movement pointed the same way in both age groups (cosine similarity 0.93 to 0.99), but sadness differed more (0.63). Once landmark noise and repeat-to-repeat variation were removed, older adults differed more from one another than young adults for anger, fear and happiness. Each 95\% interval excluded zero, although none of the three survived Holm correction (p = 0.07). Disgust showed no age difference, and for sadness it was the young adults who varied more (Holm-corrected p = 0.015). Variation among older adults may add to the difficulty for some expressions, then, but it accounts for only part of the pattern.

\begin{table}[H]
    \caption{Age gap by expression (older minus young, percentage points) for human raters and for the models, using the six-way score}
    \centering
    \begin{tabular}{lccc}
        \toprule
        \textbf{Expression} &
        \textbf{People} &
        \textbf{AffectNet-7 models} &
        \textbf{RAF-DB models} \\
        \midrule
        Neutrality       & $-11.3$ & $-60.7$ & $-9.7$  \\
        Happiness        & $-1.1$  & $0.0$   & $-0.3$  \\
        Sadness          & $-12.8$ & $-10.5$ & $-1.3$  \\
        Anger            & $-11.4$ & $+2.2$  & $-18.7$ \\
        Fear             & $-1.9$  & $0.0$   & $-37.3$ \\
        Disgust          & $-13.1$ & $-1.3$  & $-30.5$ \\
        \midrule
        All expressions  & $-8.6$  & $-11.7$ & $-16.3$ \\
        \bottomrule
    \end{tabular}
    \label{tab:expression-gap}
\end{table}

\subsection{Most of the gap lies in the models}

The observed discrepancy in model performance between younger and older faces motivates an investigation into its causes, specifically whether the issue comes from the training-data distribution or the model architecture.

The models' gaps exceeded the human gap and were far larger for particular expressions (Table~\ref{tab:expression-gap}). On older neutral faces, the AffectNet models lost about 61 points compared with 11 for people. On older fearful faces, the RAF-DB models lost 37 points and people lost 2. The AffectNet models misread 56\% of the older neutral faces that at least 90\% of raters identified correctly, and the RAF-DB models misread 52\% to 74\% of the older anger, disgust and fear faces that at least 90\% of raters identified correctly. Differences in how hard people found the photographs accounted for 29\% of the AffectNet models' age effect on neutral faces (95\% CI 15\% to 44\%). They accounted for almost none of the RAF-DB models' effect on anger, disgust and fear ($-$1\% to 8\%, with intervals that include zero).

Refitting only the final layer largely removed the age gap (Table~\ref{tab:tab4}). With each network frozen, a classifier trained on older faces was 6.1 to 13.7 points more accurate on held-out older people than one trained on young faces. The interval across people excluded zero for every model (Holm-adjusted p = 0.0002). A classifier trained on half older and half young faces recovered 76\% to 94\% of that gain. The classifier trained on older faces was about as accurate on older faces as on young ones, with gaps from $-$4.1 to +1.2 points, while the one trained on young faces left gaps of 11 to 21 points. The frozen features, then, already contain what is needed to read older faces, and the deficit sits in the final decision layer. The size of the gain depended on how strongly the classifier was regularized, ranging from 5 to 15 points across the settings tried (least for NLA), but every model gained at every setting. This experiment trains and tests within FACES, so it shows that the features are adequate. Why the released final layers read older faces poorly is a separate question.

\begin{table}[H]
    \caption{Accuracy (\%) of a linear classifier trained on frozen
    features from 36 FACES participants (young, half young and half
    older, or older) and tested on held-out older and young
    participants; means over 20 paired splits. Gain is older-trained
    minus young-trained accuracy on held-out older people, with a
    95\% bootstrap interval across people. The p-values are
    Holm-adjusted over the five models.}

    \centering
    \small
    \setlength{\tabcolsep}{4pt}
    \renewcommand{\arraystretch}{1.3}

 \begin{tabular}{
    >{\centering\arraybackslash}b{4.2cm}
    >{\centering\arraybackslash}b{1.3cm}
    >{\centering\arraybackslash}b{1.3cm}
    >{\centering\arraybackslash}b{1.4cm}
    >{\centering\arraybackslash}b{1.5cm}
    >{\centering\arraybackslash}b{3.7cm}
    >{\centering\arraybackslash}b{1.1cm}
}

    \toprule

    \makecell[b]{\textbf{Model}} &
    \makecell[b]{\textbf{Young}\\\textbf{Trained}\\\textbf{(Older)}} &
    \makecell[b]{\textbf{Young}\\\textbf{Trained}\\\textbf{(Young)}} &
    \makecell[b]{\textbf{Balanced}\\\textbf{(Older)}} &
    \makecell[b]{\textbf{Older}\\\textbf{Trained}\\\textbf{(Older)}} &
    \makecell[b]{\textbf{Gain [95\% CI]}} &
    \makecell[b]{\textbf{p}\\\textbf{(Holm)}} \\

    \midrule

    QCS (AffectNet-7)
    & 87.8 & 98.5 & 93.5 & 93.9
    & $+6.1$ {\scriptsize [+3.8, +8.3]} & 0.0002 \\

    POSTER++ (AffectNet-7)
    & 86.6 & 97.6 & 93.3 & 93.7
    & $+7.1$ {\scriptsize [+5.1, +9.8]} & 0.0002 \\

    QCS (RAF-DB)
    & 78.2 & 96.6 & 86.5 & 89.1
    & $+10.9$ {\scriptsize [+8.3, +13.6]} & 0.0002 \\

    POSTER++ (RAF-DB)
    & 73.7 & 94.7 & 84.7 & 87.4
    & $+13.7$ {\scriptsize [+11.4, +16.0]} & 0.0002 \\

    NLA (RAF-DB)
    & 74.5 & 94.2 & 81.3 & 83.2
    & $+8.8$ {\scriptsize [+5.8, +12.8]} & 0.0002 \\

    \bottomrule

\end{tabular}

    \label{tab:tab4}
\end{table}

\subsection{Alternative explanations for the age gap}
The previous section concerned model training. The focus now shifts to how other factors, such as architecture, model size, training-image quality, and pre-processing, contribute to the results. Table~\ref{tab:tab5} summarizes the control analyses. The age gap survived all four.

\begin{table}[H]
    \caption{Control analyses.}
    \centering
    \renewcommand{\arraystretch}{1}
    \begin{tabularx}{\columnwidth}{
        >{\raggedright\arraybackslash}p{0.24\columnwidth}
        >{\raggedright\arraybackslash}X
        >{\raggedright\arraybackslash}X}
        \toprule
      \multicolumn{1}{c}{\textbf{Explanation}} &
\multicolumn{1}{c}{\textbf{Test}} &
\multicolumn{1}{c}{\textbf{Result}} \\
        \midrule

        Architecture and size &
        Age gaps of the three RAF-DB models
        (11.7M to 82.9M parameters) &
        Seven-way score: gaps differ by at most
        1.5 points (Holm $p = 1.00$).
        Six-way score: by up to 4.6 points
        (Holm $p = 0.07$). \\
        \midrule

        Image quality &
        Age effect adjusted for lighting and
        exposure; age gap within thirds of
        each quality measure &
        Odds ratio for an error on older faces
        1.8--3.2 before and 1.5--3.4 after
        adjustment; gap of 9 to 20 points
        in every third. \\
        \midrule

        Face detection and cropping &
        Detection rate; age gap at crop margins
        0.0 to 0.3 &
        Detection 100\% (2,052 of 2,052);
        pooled gap of 12 to 14 points at every
        margin, each interval excluding zero. \\
        \midrule

        Expression intensity &
        Error rate in six intensity bins
        matched across age groups &
        Older faces misread 1.3 to 2.4 times
        as often in every bin; adjusting for
        intensity did not reduce the age
        odds ratio (1.9 before, 2.5 after). \\

        \bottomrule
    \end{tabularx}
    \label{tab:tab5}
\end{table}

Crop framing mattered more than age for the RAF-DB models. Some of what looks like a fixed weakness in these models, then, is really a sensitivity to preprocessing choices, a separate and more fixable issue than an inability to read certain faces at all. Widening the margin from 0.0 to 0.3 lowered their accuracy on young faces by 26 points (NLA), 34 (POSTER++) and 44 (QCS), and their recall of happiness fell to 62\% for NLA and to 0\% for both POSTER++ and QCS. The AffectNet models moved by about 5 points. The age gap itself stayed between 12.0 and 14.2 points, pooled over the five models, at every margin.

The training dataset did not change the size of the gap either. In the 2 $\times$ 2 design (QCS and POSTER++, each trained on RAF-DB and on AffectNet-7), none of the three effects on the seven-way score was significant. The training dataset shifted the gap by $-$1.6 points (95\% CI $-$7.0 to +3.3), architecture by $-$1.8 ($-$4.3 to +0.8) and their interaction by +0.5 ($-$4.0 to +4.9). The six-way score gave the same picture for the training dataset ($-$4.0, $-$8.8 to +0.8) and for architecture (+1.6, $-$0.8 to +4.0), but the interaction was significant (+6.1 points, 1.4 to 10.7; Holm p = 0.03). Within the RAF-DB models, QCS's gap was 4.6 points smaller than POSTER++'s (0.6 to 8.8; Holm p = 0.07), while on AffectNet-7 the difference was 1.4 points in the other direction.

The standard blur measure, variance of the Laplacian, behaved unexpectedly. It was 90\% higher on older faces (median 245 compared with 129). Wrinkles are fine detail, so on older faces the measure tracks skin texture rather than focus, and it was left out of the image-quality adjustment. Including a measure that partly tracks age itself would have removed some of the very effect the quality adjustment is meant to test for.

\section{Discussion}

All five models were less accurate on older adults. Because architecture and training dataset were varied independently, the gap is unlikely to reflect a single model family or dataset.

The results point to two sources of difficulty. Some of it comes from the faces themselves. Human raters and a geometry-only classifier both found older faces harder, by about 9 and 8 points, and for anger, fear and happiness older adults varied more from one another when posing. Most of it comes from the models, which lost up to 61 points on expressions where people lost 11 or fewer. The way the models failed depended on the training data. Each model's blind spot lines up with a gap in what it was trained on, rather than a single common failure mode: AffectNet models misread older neutral faces, and RAF-DB models lost fear, disgust and anger, as if each dataset had taught its models expression cues that transfer poorly to older faces.

The refitting experiment places the deficit in the final decision layer. Using the same frozen features, classifiers trained on older faces read them as accurately as younger faces. Older adults are scarce in RAF-DB, and the posed studio photographs in FACES also differ from the internet photographs in both training datasets\cite{ref19}. Either could leave the final layer poorly matched to older faces, and telling them apart needs tests on the training datasets themselves. It is possible in principle that refitting the final layer separately on each source's own images, with controlled age mixes, would show whether the deficit lies in the released weights or in what the training data contains. But that needs direct access to RAF-DB's and AffectNet's raw images and labels, which require a signed research license (AffectNet) or a direct request to the authors (RAF-DB), and cannot be redistributed either way, so it is outside the scope of this paper.

For anyone deploying FER with older adults, training data made a bigger difference than model size. For the two architectures tested on both datasets, switching the training data changed accuracy on older adults by 15 to 17 points, while switching architecture within the same training data changed it by under 2 points. The AffectNet models were more accurate and better calibrated on older faces than RAF-DB models of the same architecture. Among the RAF-DB models, the gap on the primary score stayed within 1.5 points from 11.7M to 82.9M parameters. Refitting the final layer on faces like those of the intended users is a cheap remedy worth testing, and confidence scores from the RAF-DB models should be treated as unreliable for older users. Researchers running age comparisons should calibrate crop framing, which moved the RAF-DB models' accuracy more than age did, and avoid blur measures that respond to wrinkles.

Some caveats apply. All five models share face-recognition pre-training on MS-Celeb-1M, which the design had to hold constant, since comparing five different FER trainings fairly requires holding everything else fixed, including the shared backbone. This also means the study cannot rule out that MS-Celeb-1M pre-training itself contributes to, or masks part of, the age gap. FACES contains posed studio photographs, and spontaneous expressions may behave differently. The human ratings record only whether raters chose the intended expression, so human and model confusions cannot be compared directly. Landmark noise was large for everyone, about a third of the size of the expression movements, which limits how finely differences in movement can be measured even though the noise was similar across ages. The geometry classifier's loss when transferring from young to older faces was small, so the geometric evidence rests mainly on its within-age gap.

Current FER models read older adults' expressions less accurately than younger adults', by 12 to 15 percentage points in the five models tested here. Older faces are harder for everyone: people and a geometry-only classifier lose accuracy on them, and for several expressions older adults pose less uniformly. The models, though, lose far more accuracy than people do, more than five times as much on the expressions where the gap is largest (a 61-point drop for the AffectNet models' neutral faces against an 11-point drop for human raters), and how they fail depends on their training data. AffectNet-trained models misread older neutral faces, and RAF-DB-trained models miss fear, disgust and anger. Architecture, image quality, cropping and expression intensity leave the gap in place. Refitting only the final layer on older faces nearly closed it, which means the learned features already represent older faces well enough and the problem lies in the final layer. Until the reason for the failure is known, FER systems used with older adults should be evaluated by age before deployment, and their confidence scores checked on older users. Future work should focus on how to incorporate these insights to build models with consistent and high performance on older faces.
\section{Acknowledgement}
This study used the FACES database (Ebner, Riediger, \& Lindenberger, 2010), developed at the Max Planck Institute for Human Development, Berlin, Germany. Access was granted under the FACES Platform Release Agreement. In accordance with this agreement, no FACES images are reproduced in this paper.

The completion of this research paper would not have been possible without the guidance of Dr. Md Masudur Rahman at Purdue University.

\bibliographystyle{unsrtnat}


\end{document}